\documentclass[10pt,twocolumn,letterpaper]{article}

\usepackage{cvpr}            
\usepackage[dvipsnames]{xcolor}

\usepackage{amssymb}
\usepackage{amsmath}
\usepackage{comment}
\usepackage{graphicx}

\usepackage{booktabs}
\usepackage{enumerate}
\usepackage{enumitem}
\usepackage{makecell}
\usepackage{tabu}
\usepackage{multirow}
\usepackage{utfsym}
\usepackage{colortbl} 
\definecolor{mygray}{gray}{.9}
\usepackage{pifont}
\usepackage{soul}

\definecolor{cvprblue}{rgb}{0.21,0.49,0.74}
\usepackage[pagebackref,breaklinks,colorlinks,citecolor=cvprblue]{hyperref}
\usepackage[accsupp]{axessibility} 

\begin{document}

\title{{Unveiling Parts Beyond Objects: \\ Towards Finer-Granularity Referring Expression Segmentation}}

\author{
  Wenxuan Wang$^{1,2,3}$\thanks{Equal technical contribution.} \quad
  Tongtian Yue$^{1,2*}$\quad
  Yisi Zhang$^{4}$\quad
  Longteng Guo$^{1}$\quad
  Xingjian He$^1$\\
  Xinlong Wang$^3$\quad
  Jing Liu$^{1,2}\thanks{Corresponding author.}$\\
  {$^1$
  Institute of Automation, Chinese Academy of Sciences (CASIA)}\\
  {$^2$ School of Artificial Intelligence, University of Chinese Academy of Sciences (UCAS)}\\
  {$^3$ Beijing Academy of Artificial Intelligence (BAAI)}\\
  {$^4$ University of Science and Technology Beijing (USTB)}\\
  \tt\small \{wangwenxuan2023@ia.ac.cn, wangxinlong@baai.ac.cn, jliu@nlpr.ia.ac.cn\}
}

\twocolumn[{
\renewcommand\twocolumn[1][]{#1}
\maketitle 
\vspace{-10mm}
\begin{center} 
\centering 
\includegraphics[width=0.82\textwidth]{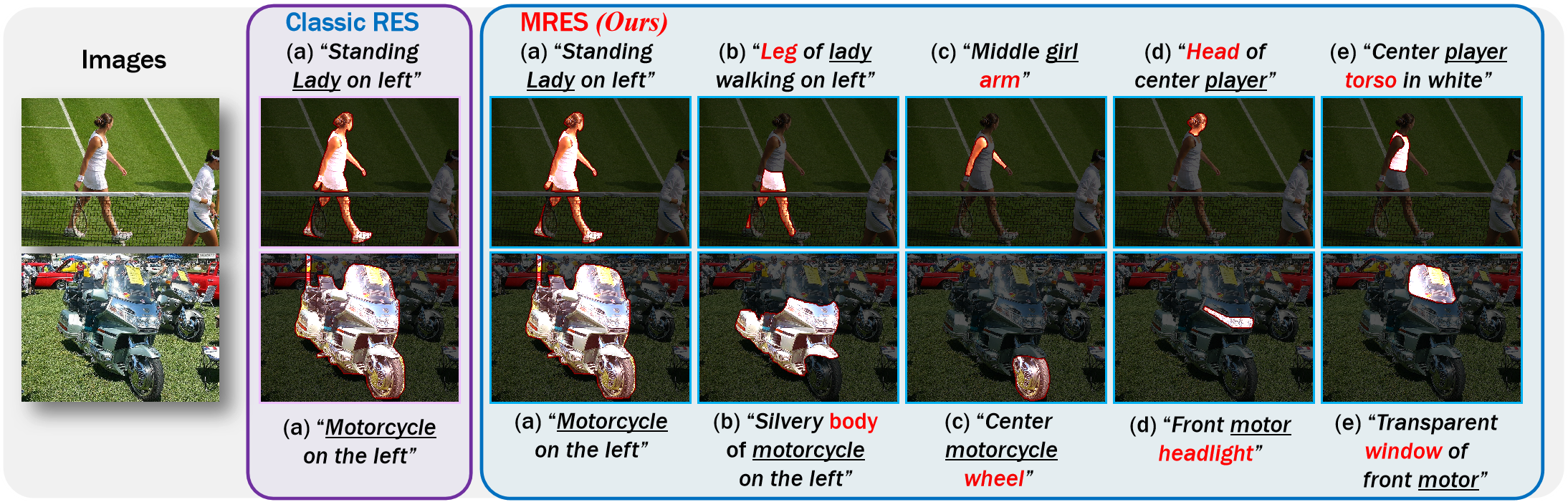} 
\vspace{-3mm}
\captionof{figure}{Classic Referring Expression Segmentation (RES) only supports expressions that indicate a single target object, \eg, (a). Compared with classic RES, the proposed \textbf{Multi-Granularity Referring Expression Segmentation (MRES)} task supports expressions indicating the specific
\textbf{\textit{part-level regions}} of target objects, \eg, part-level expressions like ({b})-({e}) from our newly built RefCOCOm benchmark.}
\label{fig:intro} 
\end{center}
}]

\renewcommand{\thefootnote}{\fnsymbol{footnote}}
\footnotetext[1]{Equal contribution.}
\footnotetext[2]{Corresponding author.}

\begin{abstract}
\vspace{-10pt}
Referring expression segmentation (RES) aims at segmenting the foreground masks of the entities that match the descriptive natural language expression. 
Previous datasets and methods for classic RES task 
heavily rely on the prior assumption that one expression must refer to object-level targets.
In this paper, we take a step further  
to finer-grained part-level RES task.
To promote the object-level RES task towards finer-grained vision-language understanding, we put forward a new multi-granularity referring expression segmentation (MRES) task and construct an evaluation benchmark called RefCOCOm by manual annotations.
By employing our automatic model-assisted data engine, 
we build the largest visual grounding dataset namely MRES-32M, which comprises over 32.2M high-quality masks and captions on the provided 1M images. 
Besides, a simple yet strong model named UniRES is designed to accomplish the unified object-level and part-level grounding task. 
Extensive experiments on our RefCOCOm for MRES and three datasets (\ie, RefCOCO(+/g)) for classic RES task demonstrate the superiority of our method over previous state-of-the-art methods.
To foster future research into fine-grained visual grounding, 
our benchmark RefCOCOm,
the MRES-32M dataset and model UniRES will be publicly available at \url{https://github.com/Rubics-Xuan/MRES}.

\end{abstract}    
\section{Introduction}
\label{introduction}

As one of the most challenging tasks in vision-language understanding, referring expression segmentation (RES) aims to locate specific regions at the pixel level based on a descriptive language expression. 
Compared to traditional visual segmentation tasks that focus on images or videos alone, RES poses greater difficulties and challenges due to the necessity of strong comprehension across modalities, 
but it can simultaneously alleviate the problem of pre-defined categories in conventional object detection or segmentation. 
With the real-world scene often requiring diversity in target identification, RES task holds vast potential for applications, \eg, language-based human-object interaction and interactive image editing.

Since the concept of RES task was initially proposed in \cite{hu2016segmentation}, various multimodal frameworks such as \cite{hu2016segmentation,wang2022cris,yang2022lavt,kim2022restr} towards precise RES have been proposed to deal with the challenging feature extraction and alignment problems between visual and linguistic modalities.
However, current works are limited to the scope of object-level grounding. 
As shown in Table \ref{tab:dataset_compare}, the RefCOCO dataset \cite{yu2016modeling} stands as one of the most widely used grounding benchmarks by far, basically containing only object-level visual and textual annotations. \textit{It does not take into account the part-level grounding task, which is crucial for future multimodal models to act as intelligent agents to realize the fine-grained perception of real world.} 
The existing grounding datasets couldn't support the training and evaluation of such advanced capabilities.
Thus, exploring how to transcend the current object-level constraints and delve into finer-grained part-level grounding, is highly meaningful and worthy of in-depth study, which is the primary focus of this work.

\begin{table}[t]
  \renewcommand\arraystretch{1.2}
  \centering
  \footnotesize
  \caption{Comparison among different referring expression datasets, including ReferIt \cite{kazemzadeh2014referitgame}, RefCOCO(+/g) \cite{yu2016modeling,nagaraja2016modeling}, PhraseCut \cite{wu2020phrasecut}, and our proposed \textbf{MRES-32M}. Part-Level: expressions that specifies various parts of the target object in the given image.}
  \vspace{-3mm}
  \setlength{\tabcolsep}{0.6mm}
  {\begin{tabular}{lcccc}
    \specialrule{.1em}{.05em}{.05em} 
          & ReferIt & RefCOCO(+/g) & PhraseCut&  \textbf{MRES-32M}\\
          \cline{2-5}
          Image Source  & CLEF \cite{grubinger2006iapr} & COCO \cite{lin2014microsoft} & VG \cite{krishna2017visual}  & Object365 \cite{shao2019objects365} \\
          Object-Level & \ding{51} & \ding{51} & \ding{51} & \ding{51} \\
          Part-Level & \ding{53} & \ding{53} & \ding{53} & \ding{51} \\
          \makecell[c]{Expression Type} & free  & free  & templated & free \\
    \specialrule{.1em}{.05em}{.05em} 
    \end{tabular}}%
  \label{tab:dataset_compare}%
\vspace{-15pt}
\end{table}%

In fact, prior to this work, a few works along another research line have made significant strides towards more fine-grained visual understanding.
The demand for finer-grained visual understanding of objects arouses research community into constructing high-quality part-level dataset. 
Specifically, recent years have witnessed the introduction of various datasets that provide fine-grained part-level masks and bounding boxes annotations for either objects of specific categories \cite{gong2017look,reddy2018carfusion,wah2011caltech} or general categories \cite{he2022partimagenet,mo2019partnet,chen2014detect,ramanathan2023paco}.
However, these datasets essentially correspond to unimodal (\ie, visual) downstream tasks, lacking a deep connection between fine-grained part-level masks and rich textual descriptions.
To our knowledge, few of previous studies have established this connection.
Consequently, there appears to be a limited availability of fine-grained, large-scale vision-language datasets that facilitate part-level cross-modality understanding, which is necessary in terms of two aspects.
Firstly, when describing an object, people always naturally gravitate towards detailing part-level local features, underscoring the indispensable need for multimodal agents to understand the part granularity. 
Secondly, a fine-grained understanding at the part level can positively promote the perception of object-level targets, especially under extreme conditions such as occlusion or deformation, which will consistently propel advancements in widely focused object-level tasks like classic visual grounding.

Therefore, in this work, we attempt to fill this important blank space that has been neglected before and move towards finer-grained part-level RES task.
Specifically, to push towards finer-grained vision-language understanding, we propose a new multi-granularity referring expression segmentation (MRES) task and an evaluation benchmark named RefCOCOm by manually annotating the part-level targets based on the previous benchmark RefCOCO with only object-level labels.
We also construct the largest-scale visual grounding dataset which is also the first dataset to support part-level visual-textual annotations, and build a simple baseline model to accomplish the unified multi-granularity (\ie, object-level and part-level) RES task.

Our main contributions can be summarized as follows:
\setlist{nolistsep}
\begin{itemize}[noitemsep,leftmargin=*]
    \item 
    We propose a new MRES task (as shown in Fig. \ref{fig:intro}) with corresponding benchmark RefCOCOm for evaluation, pushing classic RES task towards finer-granularity understanding of real-world scenes.
    \item 
    We build a multi-grained visual grounding dataset namely MRES-32M, which to the best of our knowledge is the first grounding dataset that supports part-level vision-language annotations and also the largest-scale visual grounding dataset.
    \item 
    To effectively unify both the object-level and part-level RES tasks, we propose a simple yet strong model namely UniRES, 
    which achieves the new state-of-the-art (SOTA) performance on three object-level benchmarks for classic RES tasks and our multi-granularity RefCOCOm benchmark for the proposed MRES task.
\end{itemize}

\section{Related Work}
\label{relatedwork}

\textbf{Referring Expression Segmentation.}
As a challenging visual grounding task  
the concept of RES is first proposed by \cite{hu2016segmentation}.
Subsequent works such as \cite{yu2018mattnet,ye2019cross,hu2020bi,huang2020referring,wang2022cris} predominantly follow a two-step pipeline of encoding the linguistic and visual features separately and deriving the fused multimodal features from unimodal representations for mask prediction.
In this way, the effectiveness of obtained multimodal representations essentially dominates the model performance, which has been continually studied in the following research \cite{ding2021vision,yang2022lavt,lai2023lisa,zou2023segment,zou2023generalized}.
Recently, there are some works \cite{yu2023zero,ni2023ref,suo2023text} focusing on zero-shot RES task which has great application potential.
In addition, 
a few recent works \cite{hu2023beyond,liu2023gres} have been devoted to address the limitations of the existing benchmark datasets \cite{kazemzadeh2014referitgame,yu2016modeling,nagaraja2016modeling,wu2020phrasecut} for visual grounding task.
However, it's worth noting that previous works primarily focus on classic object-level RES methods and datasets, paying few attention to the vital fine-grained part-level grounding.

\noindent \textbf{Part-level Visual Perception.} 
The increasing interest in the fine-grained understanding of objects has driven the creation of part-level annotated datasets across both specialized and general domains. 
Pioneering research in the former field has introduced datasets with part-level annotations, focusing on objects of particular categories like human body parts \cite{gong2017look}, animal body parts \cite{reddy2018carfusion} and vehicle components \cite{wah2011caltech}. In contrast, more general datasets that offer part annotations for a variety of common objects include Pascal-Part \cite{chen2014detect}, PartNet \cite{mo2019partnet}, and PartImageNet \cite{he2022partimagenet}. 
Furthermore, recent work \cite{de2021part} introduces a finer-grained dataset with additional instance-specific part annotations for panoptic segmentation task. 
PACO \cite{ramanathan2023paco} extends the available object-level datasets by including the annotations of part granularity and attribute for finer-grained segmentation and detection tasks. 
Moreover, VLPart \cite{sun2023going} proposes a parsing pipeline for part segmentation and a detector that is capable of realizing the segmentation task of both open-vocabulary objects and their part regions.

\section{Multi-Granularity Grounding Benchmark}
\label{Multi-Granularity Visual Grounding Benchmark RefCOCOm}

\subsection{Multi-Granularity RES Task}

The interaction capability of a multimodal agent directly depends on the visual granularity it can perceive and understand. However, in the current research landscape, multi-granularity visual grounding remains under-explored. To break the limitations wherein visual-language alignment is constrained to the object level, we intend to take a step forward towards finer-grained vision-language understanding and propose the multi-granularity RES task, which requires models' proficiency in uniformly grounding entities across both part and object levels in response to various textual references.
Within the single modality, it entails an enhancement of the capacity to apprehend complex, implicit knowledge structures, \eg, the granularity delineated by textual descriptions and the advanced level of pixel-wise semantic comprehension. 
Across the vision-language modalities, finer-grained references also pose new challenges for the accuracy and robustness of cross-modal alignment.
Besides, since previous studies are mainly restricted to object-level referring segmentation and related benchmarks for the evaluation of part-level grounding performance are unavailable, we build a corresponding benchmark named RefCOCOm for performance evaluation, which will be elaborated below.

\subsection{RefCOCOm Benchmark}

\noindent \textbf{Manual Annotation Pipeline.}
\textit{To facilitate easy evaluation by community researchers without the need for any code modifications and build the multi-granularity benchmark with excellent quality, we split the existing object masks into part-level annotations through manual labeling on three subsets (\ie, validation, testA and testB) of the most prevalent RefCOCO \cite{yu2016modeling} dataset.}
Initially, we establish a set of criteria for part masks delineation and identify the specific part categories requisite for each object. Subsequently, we employ SAM \cite{kirillov2023segment} to procure the preliminary segmentation results. We recruit 30 skilled annotators to refine and captioning these results through an online annotation tool developed by ourselves. 
To ensure the annotation quality, we conduct spot checks at three progress nodes of 20\%, 50\%, and 80\% of the total workload with corrective feedback.

\begin{figure}[thbp]
    \centering
    \includegraphics[width=0.46\textwidth]{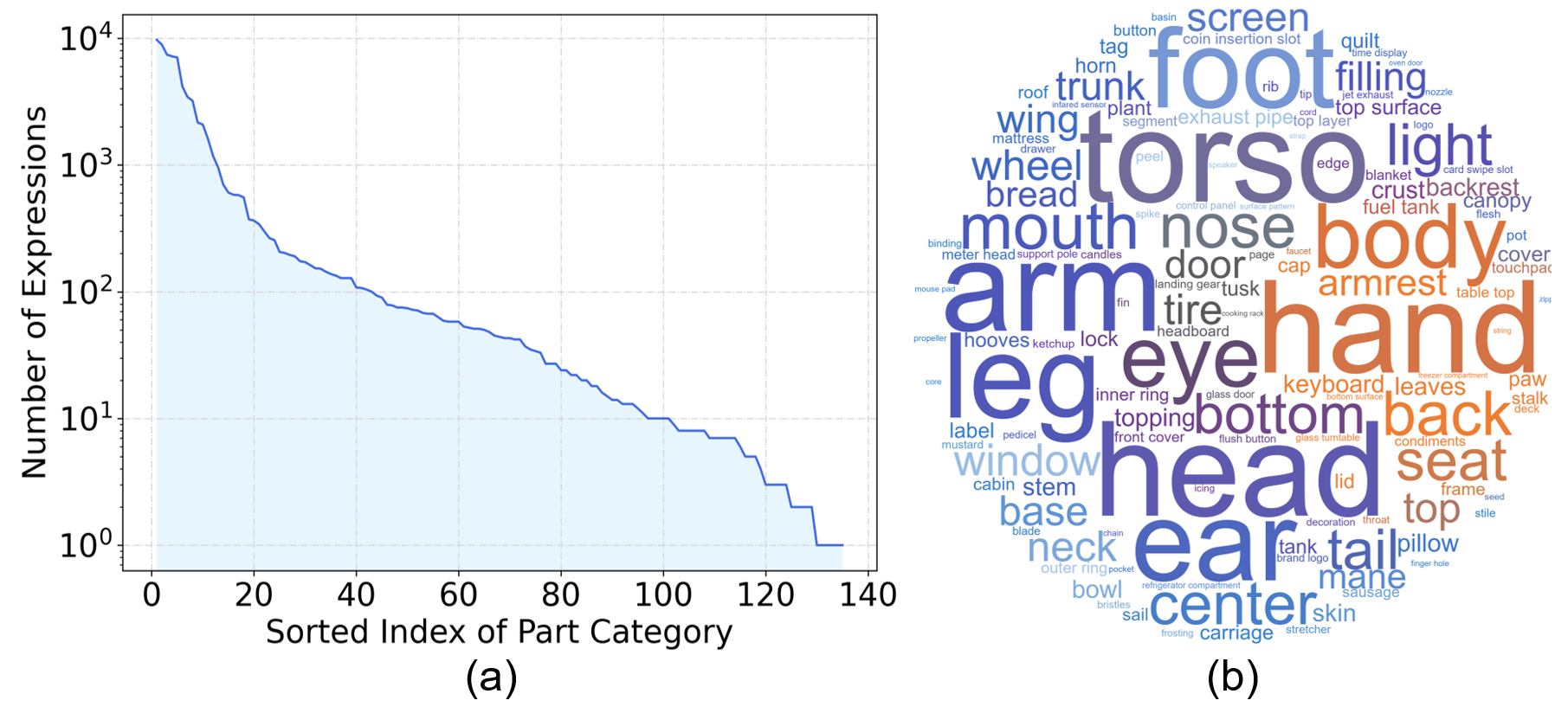}
    \vspace{-10pt}
    \caption{RefCOCOm benchmark statistics.
    (a) the number of referring expressions per parts' category in the log scale. 
    (b) the word cloud highlights the head categories.
    }
    \label{fig_RefCOCOm}
    \vspace{-15pt}
\end{figure}

\noindent \textbf{Benchmark Details.}
Totally, we collect 70k part references and corresponding 26k masks. 
The average length of the references is 5.1, covering 80 object and 391 part categories, among which the number of referring expressions per parts’ category and the word cloud that highlights the head categories are both presented in Fig. \ref{fig_RefCOCOm}.
Mixed with the original object-level annotations, our RefCOCOm has a total of 34k masks and 92k references. 
As an enhanced expansion of RefCOCO \cite{yu2016modeling}, our RefCOCOm puts forward higher requirements for referential understanding and visual perception capabilities. 
To better match the intention of multi-granularity unification, we leverage mean Intersection-over-Union (mIoU) as evaluation metrics.

\begin{figure*}[htbp]
    \centering
    \includegraphics[width=0.82\textwidth]{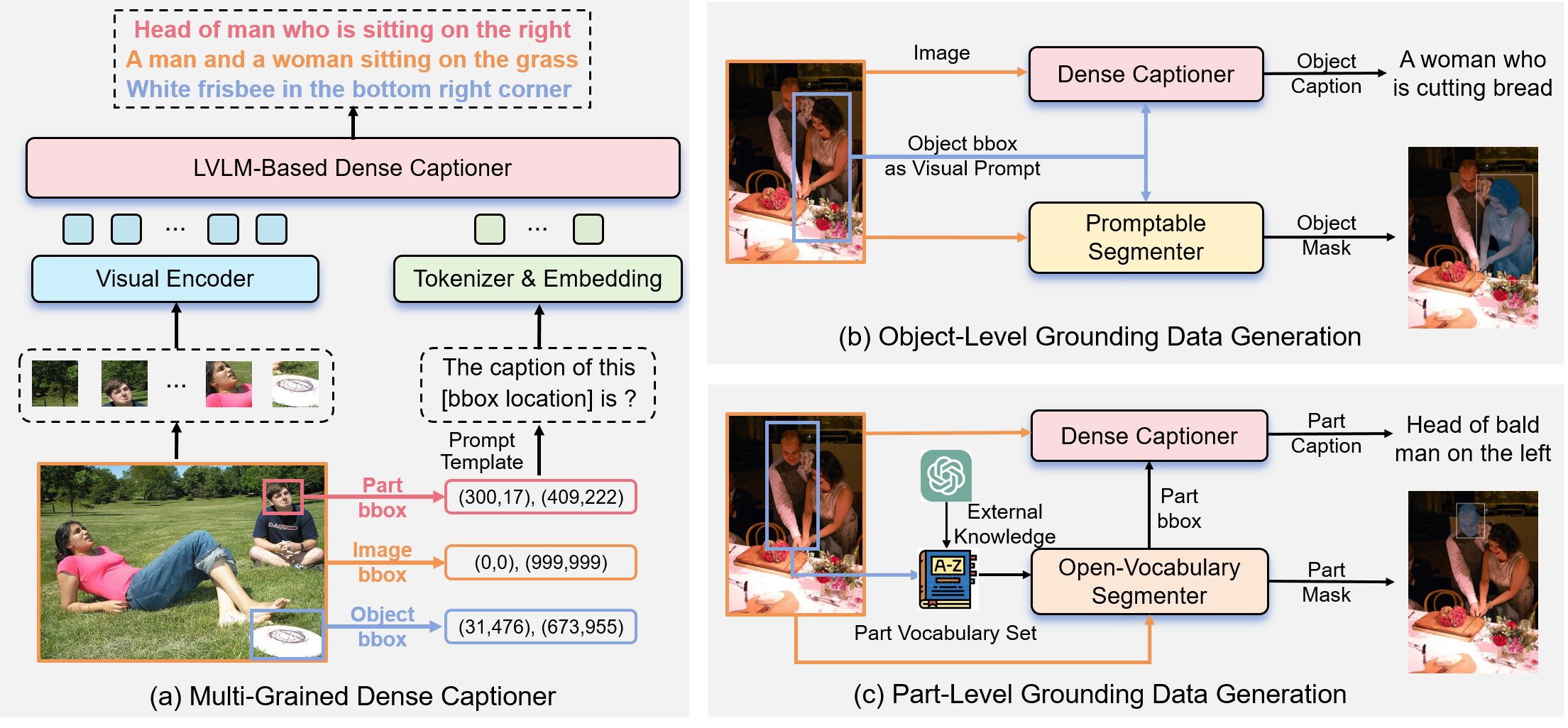}
    \vspace{-8pt}
    \caption{
    The illustration of our data engine for building the MRES-32M dataset. 
    (a) We start by fine-tuning an LVLM to create a capable dense captioner, which can effectively handle captioning tasks at three levels of granularity.  
    (b) To generate object-level grounding data, we feed images and original bounding boxes into the dense captioner and a powerful segmenter to obtain the captions and masks of various objects. 
    (c) We leverage the external knowledge from LLMs to decompose the existing object category annotations into a vocabulary set of part-level tags, which are sequentially fed into an open-vocabulary segmenter and our captioner to acquire the part-level annotations.
    }
    \label{fig_dataengine}
    \vspace{-12pt}
\end{figure*}

\section{Multi-Granularity Grounding Dataset}
\label{Multi-Granularity Visual Grounding Dataset MRES-32M}

\subsection{Data Collection Engine}

Given the intrinsic complexity of classic RES task, the associated training data necessitates extensive annotation costs across both textual and visual domains. 
When the granularity is promoted to part-level, the annotation intricacies are further exacerbated. 
We contend that the primary bottleneck hindering the emergence of open-world grounding is the limitation imposed by the current scalability of data. By leveraging robust foundation models for synergistic enhancement, we introduce an advanced data engine capable of automatically generating reliable visual grounding data.

\noindent \textbf{Multi-Grained Dense Captioner.} 
As the large language models (LLMs) \cite{chiang2023vicuna, touvron2023llama} continue to empower the multimodal domain with profound capabilities, the current scope of large vision-language models (LVLMs) \cite{bai2023qwen, chen2023minigpt, chen2023shikra} has effectively expanded into the grounding domain. Formally, these models utilize numerical coordinates to represent bounding boxes, allowing positional information to be seamlessly integrated into the language model via word embeddings for contextual understanding. However, the research community's exploration into the part granularity remains nascent. The constrained training data have resulted in the fine-grained descriptive capabilities of LVLMs being confined merely to the object-level. To fully harness the open-world visual knowledge acquired from extensive pretraining on image-level and object-level data, we have devised a unified fine-tuning scheme that equips LVLMs to operate as dense captioners across all granularity levels, as depicted in Fig. \ref{fig_dataengine} (a). 
Data for all three granularity levels are sourced from manual annotations to ensure reliability. The input comprises an image and the corresponding bounding box to be described, with bounding box coordinates normalized to integer values within the [0,999] range. 
For image granularity, we employ the COCO dataset \cite{lin2014microsoft}, where all bboxes are uniformly represented as $(0,0),(999,999)$. 
For object granularity, we utilize the Visual Genome dataset \cite{krishna2017visual}. 
For part granularity, we draw upon unimodal semantic segmentation data \cite{ramanathan2023paco, he2022partimagenet, chen2014detect} and employ a template in the form of $PartNameX\; of\; ObjectNameY$ to construct dense captions. 
This unified multitask training approach can be synergistic across different granularity: it allows LVLMs to incorporate more comprehensive and detailed information to enhance part granularity descriptions. Concurrently, knowledge of part granularity assists LVLMs in generalizing knowledge within object interiors.

\vspace{-0.5mm}
\noindent \textbf{Model-Assisted Data Generation.} 
For the data generation of object-level visual grounding, we capitalize on large-scale object detection dataset Object365 \cite{shao2019objects365} to furnish highly reliable bounding boxes. Moreover, the rich category labels inherent in it ensure a comprehensive knowledge coverage. As illustrated in Fig. \ref{fig_dataengine} (b), the bounding boxes will serve as visual prompts, which are independently sent into both a promptable segmenter (\ie segment anything model \cite{kirillov2023segment}) and our dense captioner to obtain the segmentation masks and detailed semantic descriptions, respectively.

For part-level grounding data, we propose a hierarchical annotation scheme that builds upon existing object-level annotations, as shown in Fig. \ref{fig_dataengine} (c). 
Specifically, employing GPT-4 \cite{openai2023gpt} endowed with extensive external knowledge, we decompose the objects present in a given image to generate a customized part vocabulary set. This tailored vocabulary set is then fed into an open-vocabulary segmenter \cite{sun2023going}, which yields precise part masks and bounding boxes. These bounding boxes are subsequently sent into our dense captioner to acquire corresponding detailed captions.

\noindent \textbf{Data Filtering.} 
After completing the multi-granularity annotation of all images, we further introduce CLIP \cite{clip} for filtering. The bounding box is cropped from the original image, and then sent to the encoder together with the dense caption to measure the similarity. 
To ensure the consistency between the visual and linguistic annotations to a great extent, we retain box-caption pairs with similarity greater than 0.5 as the final annotation results.

\subsection{MRES-32M Dataset Details}

\begin{table}[thbp]
    \centering
    \small
    \setlength{\tabcolsep}{2.25pt} 
    \caption{Comparisons with previous object-level visual grounding datasets and part-level segmentation datasets. \# denotes the specific number, 
    where Cats and Avg Len denote the object/part categories and the average length of referring expressions.
    ``-'' denotes the corresponding part masks or captions are unavailable. }
    \vspace{-3mm}
    \begin{tabular}{lccccc}
    \specialrule{.1em}{.05em}{.05em}
        Dataset & \#Imgs & \#Objs & \#Parts & \#Cats & \#Avg Len \\
        \midrule
        \multicolumn{6}{l}{\color{gray} Object-Level Visual Grouding} \\
        ReferIt \cite{kazemzadeh2014referitgame} & 20K & 97K & -- & 238/-- & 3.2 \\
        RefCOCO \cite{yu2016modeling} & 20K & 50K & -- & 80/-- & 3.6 \\
        RefCOCO+ \cite{yu2016modeling} & 20K & 49K & -- & 80/-- & 3.5 \\
        RefCOCOg \cite{nagaraja2016modeling} & 26K & 54K & -- & 80/-- & 8.4 \\
        GRES \cite{liu2023gres}& 20K & 60K & -- & 80/-- & 3.7 \\
        \midrule
        \multicolumn{6}{l}{\color{gray} Part-Level Segmentation \& Detection} \\
        PartsIN \cite{he2022partimagenet}  & 24K & 24K & 112K & 158/609 & -- \\
        PascalPart \cite{chen2014detect} & 19K & 40K & 363K & 20/193 & -- \\
        PACO \cite{ramanathan2023paco} & 20K & 260K & 641K & 75/456 & -- \\
        \midrule
        \multicolumn{6}{l}{\color{gray} Multi-Grained Visual Grounding} \\
        \rowcolor{mygray}
        MRES-32M (Ours) & \textbf{1M} & \textbf{15.3M} & \textbf{16.9M} & \textbf{365/2299} & 4.6 \\
    \specialrule{.1em}{.05em}{.05em}
    \end{tabular}
    \label{tab:dataset}
    \vspace{-5mm}
\end{table}

As listed in Table \ref{tab:dataset}, existing datasets, such as the most commonly used benchmark dataset RefCOCO \cite{yu2016modeling}, have limitations in terms of small data scale and lacking part-level dense annotations. As one of the pioneering works for finer-granularity unimodal visual comprehension, PACO \cite{ramanathan2023paco} only contains semantic tags of objects or parts, without informative descriptions and visual context encompassed as referring expressions. 
Summarily, we compare the proposed MRES-32M with existing datasets and list some unique and significant properties of our dataset in Table \ref{tab:dataset}.
Besides, we have also provided a few examples in our MRES-32M dataset in the appendix.

\noindent \textbf{Unified Multi-Granularity.}
In comparison to the grounding counterparts, our MRES-32M is the first visual grounding dataset covering both part and object granularity. In comparison to the part-level segmentation counterparts, our MRES-32M provides informative and unique fine-grained description for each part mask.\\
\noindent \textbf{More Diversified Categories.}
Our MRES-32M is composed of 365 object categories and an associated 2,299 part categories. Compared with existing datasets, it covers a wider range of multimodal knowledge and is an important step towards open-world understanding.\\
\noindent \textbf{Breakable Data Scales.}
To the best of our knowledge, MRES-32M is the largest dataset in the current grounding research community. In terms of the number of images and object instances, it surpasses the largest existing visual grounding dataset RefCOCOg \cite{nagaraja2016modeling} by factors of 38 and 283, respectively. Meanwhile, it encompasses part instance counts that exceed the largest existing part semantic segmentation dataset \cite{ramanathan2023paco} by 58 times.\\
\noindent \textbf{More Complex References.}
Benefiting from our LVLM-based dense captioner, the reference of MRES-32M could be more fully combined with the visual context for entity (\ie, part and object) description. Without sticking to a specific template, the relationships and attributes of entities could be highlighted in free natural language expressions.

\begin{figure*}[thbp]
    \centering
    \includegraphics[width=0.82\textwidth]{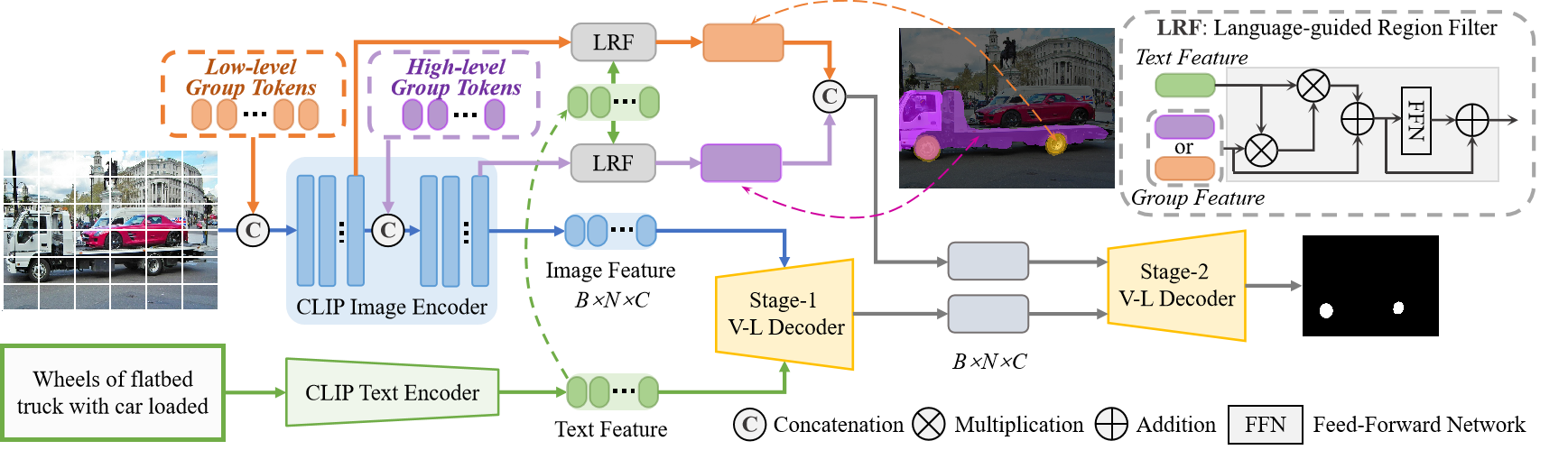}
    \vspace{-10pt}
    \caption{
    The architecture of our UniRES model as a simple baseline for the MRES task. UniRES mainly comprises three parts: the visual and textual backbone for feature extraction, the pixel grouping design for aggregating the low-level and high-level features, and the cascaded two-stage vision-language decoder for multimodal feature fusion and the generation of segmentation masks.
    }
    \label{fig_model}
    \vspace{-15pt}
\end{figure*}

\section{Multi-Granularity RES Model}
\label{methodology}

Next, we describe the proposed multi-granularity RES model UniRES for the unified MRES task with the referring targets at both object and part granularity. 
\textit{Since our original intention is to establish a simple and easy-to-follow baseline model for the proposed multi-granularity RES task, the structure of our model UniRES is designed to be simple and clean.}
As shown in Fig. \ref{fig_model}, UniRES has three major components, which will be illustrated below.

\noindent \textbf{Vision-Language Backbone.} 
Taking account of the required attributes of both strong ability to capture the vision-language feature representations and promising scalability, we leverage the CLIP pre-trained weights of CLIP model \cite{radford2021learning}, which learns transferable visual and linguistic concepts from vast image-text pairs, and adopt it as the backbone for our referring segmentation framework. The respective image and text encoders (\ie Vision Transformer (ViT) \cite{dosovitskiy2020image} and Transformer \cite{vaswani2017attention} respectively) from CLIP are utilized to effectively extract visual and linguistic features.

\noindent \textbf{Query-based Grouping Design.}
To effectively accomplish the proposed MRES task, it is essential to exploit both low-level local and high-level global visual features. 
In order to enhance the local-global visual representations of CLIP backbone without introducing much additional parameters or altering model's structure, we have incorporated 64 and 8 learnable tokens (set empirically) into the first and middle layers of CLIP visual backbone. These learnable tokens traverse the first and second halves of the visual backbone. 
We expect that the ViT's internal self-attention mechanism implicitly serves as a manner to perform visual grouping, obtaining representative group tokens that capture the low-level local and high-level global features simultaneously. Based on the fact that local features spread more fragmented, the number of appended low-level group tokens is greater than that of high-level counterparts.
Then, group tokens from both levels are fed into a language-guided region filter (LRF) to select the language-related visual features by cross-attention mechanism, which is followed by concatenation to fuse these expression-related visual group tokens for subsequent vision-language decoding.

\noindent \textbf{Two-Stage V-L Decoder.}
Now that the visual and textual feature representations from backbone, and the expression-related group tokens with two different levels are obtained, the two-stage mask decoder which comprises stacked Transformer layers is utilized to generate the segmentation masks. 
Specifically, the first-stage V-L decoder takes the extracted visual and textual features as input and generates first-stage fused multimodal representations. 
Subsequently, these multimodal features are further integrated with the grouped expression-related region features of the two semantic levels (\ie, low-level \& high-level) to realize further feature enhancement, which is followed by a linear projection layer to obtain the final segmentation masks.

\section{Experimental Results}
\label{experiment}

To evaluate the effectiveness of our method, comprehensive experiments are conducted on three classic RES datasets (\ie, RefCOCO \cite{yu2016modeling}, RefCOCO+ \cite{yu2016modeling}, RefCOCOg \cite{nagaraja2016modeling}) and our RefCOCOm benchmark for multi-granularity RES.

\subsection{Datasets}

The details about the three benchmark datasets for classic RES task can be found in the appendix.

\subsection{Experimental Setup}

\noindent \textbf{Implementation Details.} 
The implementation details are provided in the appendix.

\noindent \textbf{Evaluation metrics.} 
We adopt mean Intersection-over-Union (mIoU) and overall Intersection-over-Union (oIoU) 
as evaluation metrics. 
The mIoU measures the ratio between the intersection area and the union area of the predictions and labels among the test samples, while the oIoU calculates the total intersection
area over total union area.

\subsection{Main Results}

\subsubsection{Multi-Granularity MRES Task}

\noindent \textbf{Comparison with SOTA Methods.}
To evaluate the multi-granularity grounding performance of our UniRES as the baseline for the proposed MRES task and previous RES methods, we further conduct experimental comparison on our newly built RefCOCOm benchmark dataset.
As presented in Table \ref{tab:results_mgres}, the classic RES methods including four specialist models (\ie, SeqTR \cite{zhu2022seqtr}, CRIS \cite{wang2022cris}, LAVT \cite{yang2022lavt})
and two generalist models (\eg, X-Decoder \cite{zou2023generalized} and SEEM \cite{zou2023segment}) are incorporated.
For fair comparisons, we re-implement these SOTA methods and report their performance on our RefCOCOm.
It is clear that either the specialist models for classic RES task or the powerful generalist models perform poorly on RefCOCOm, which requires the crucial skills of both part-level and object-level referring segmentation.
Benefiting from our MRES-32M dataset, our UniRES can better master the part-level RES skills and handle the multi-granularity RES task, achieving considerably higher segmentation accuracy.
Due to the significantly higher difficulty of part-level RES (\ie multi-granularity RES) compared to classic RES, the absolute value of segmentation accuracy is correspondingly lower. 
This further emphasizes the importance of researching finer-grained part grounding where previous SOTA methods have fallen short.

\begin{table}[htbp]
  \renewcommand\arraystretch{1.05}
  \centering
  \footnotesize
  \caption{Comparison with previous SOTA methods on our RefCOCOm benchmark in terms of mIoU. Part and Obj\!\;\&\;\!Part denote part-only and multi-grained evaluation settings of our MRES task.}
  \vspace{-3mm}
  \centering
     \setlength{\tabcolsep}{1.0mm}{\begin{tabular}{l|cc|cc|cc}
      \specialrule{.1em}{.05em}{.05em} 
      \multirow{2}{*}{Methods} & \multicolumn{2}{c|}{val} & \multicolumn{2}{c|}{testA} & \multicolumn{2}{c}{testB} \\
       & Part & Obj\!\;\&\;\!Part  & Part & Obj\!\;\&\;\!Part & Part & Obj\!\;\&\;\!Part  \\
        \hline
        \textit{Specialists} & \multicolumn{6}{l}{} \\
        \hline
        SeqTR~\cite{zhu2022seqtr} & 13.9 & 28.2 & 12.1 & 22.8 & 18.1 & 34.7 \\
        CRIS~\cite{wang2022cris} & 10.6 & 25.4 & 10.1 & 21.2 & 12.9 & 30.0\\
        LAVT~\cite{yang2022lavt} & 15.3  & 29.9 & 13.2 & 24.4 & 18.7 & 35.5\\
        \hline
        \textit{Generalists} & \multicolumn{6}{l}{} \\
        \hline
        X-Decoder~\cite{zou2023generalized} &  16.2 & 29.5 &  13.6 & 23.6 & 20.3 & 33.8\\
        SEEM~\cite{zou2023segment} & 16.1 & 29.4 & 13.6 & 23.4 & 20.4 & 33.9\\
        \hline
        \rowcolor{mygray}\textbf{UniRES} (Ours) & \textbf{19.6} & \textbf{34.3} & \textbf{16.4} & \textbf{27.8} & \textbf{25.2} & \textbf{41.7}\\
        \specialrule{.1em}{.05em}{.05em} 
\end{tabular}}
\vspace{-5mm}
\label{tab:results_mgres}
\end{table}

\noindent \textbf{Qualitative Analysis.} 
We also conduct visual comparisons of CRIS \cite{wang2022cris}, LAVT \cite{yang2022lavt} and our UniRES on the MRES task, which is provided in appendix. 

\vspace{-2mm}
\subsubsection{Classic Object-Level RES Task}

\begin{table*}[htbp]
    \small
    \setlength{\belowcaptionskip}{1.0pt}
    \begin{center}
    \caption{Comparisons with the state-of-the-art approaches on previous three classic RES benchmark datasets under both the zero-shot and fine-tuning settings.
    ``-'' denotes that the result is not provided.}
    \vspace{-3mm}
    \setlength{\tabcolsep}{2.0mm}{
    \begin{tabular}{l|l|ccc|ccc|cc}
    \specialrule{.1em}{.05em}{.05em}
        \multicolumn{2}{c|}{\multirow{2}{*}{Method}} &  \multicolumn{3}{c|}{RefCOCO} & \multicolumn{3}{c|}{RefCOCO+} & \multicolumn{2}{c}{RefCOCOg} \\
        \multicolumn{2}{c|}{} & val & testA & testB & val & testA & testB & val & test \\
        \specialrule{.1em}{.05em}{.05em}
        & \multicolumn{1}{l|}{\textit{Zero-Shot Methods}} & \multicolumn{8}{l}{} \\
        \midrule
        \multirow{7}{*}{\rotatebox{90}{mIoU}} & Region token~\cite{yu2023zero} \textcolor{gray}{(CVPR-23)} & 23.4  & 22.1 & 24.6  & 24.5 & 22.6 & 25.4 & 27.6 & 27.3 \\
        & Cropping~\cite{yu2023zero} \textcolor{gray}{(CVPR-23)} & 24.8 & 22.6 & 25.7 & 26.3 & 24.1 & 26.5 & 31.9 & 30.9  \\ 
        & Global-Local CLIP~\cite{yu2023zero} \textcolor{gray}{(CVPR-23)}  & 26.2 & 24.9  & 26.6 & 27.8 & 25.6 & 27.8 & 33.5 & 33.7  \\
        & SAM-CLIP~\cite{ni2023ref} \textcolor{gray}{(arXiv-23)} & 26.3 & 25.8 & 26.4 & 25.7 & 28.0 & 26.8 & 38.8 & 38.9  \\
        & Ref-Diff~\cite{ni2023ref} \textcolor{gray}{(arXiv-23)} & {37.2} & {38.4} & {37.2} & {37.3} & {40.5} & {33.0} & {44.0} & {44.5}  \\
        & TAS~\cite{suo2023text} \textcolor{gray}{(arXiv-23)} & {39.8} & {41.1} & {36.2} & {43.6} & {49.1} & {36.5} & {46.6} & {46.8}  \\
        \cline{2-10}
        \rowcolor{mygray} \cellcolor{white} &  \textbf{UniRES} (Ours) & \textbf{71.2} & \textbf{74.8} & \textbf{66.0} & \textbf{59.9} & \textbf{66.7} & \textbf{51.4} & \textbf{62.3} & \textbf{63.2} \\
        \midrule
        & \multicolumn{1}{l|}{\textit{Fine-Tune Methods}} & \multicolumn{8}{l}{} \\
        \midrule
        \multirow{8}{*}{\rotatebox{90}{oIoU}} & EFNet \cite{feng2021encoder} \textcolor{gray}{(CVPR-21)} & 62.8 & 65.7 & 59.7 & 51.5 & 55.2 & 43.0 & - & - \\
        & LTS \cite{jing2021locate} \textcolor{gray}{(CVPR-21)}  & 65.4 & 67.8 & 63.1 & 54.2 & 58.3 & 48.0 & 54.4 & 54.3 \\
        & ReSTR \cite{kim2022restr} \textcolor{gray}{(CVPR-22)}  & 67.2 & 69.3 & 64.5 & 55.8 & 60.4 & 48.3 & - & - \\
        & ReLA~\cite{liu2023gres} \textcolor{gray}{(CVPR-23)} & 73.8 & 76.5 & 70.2 & 66.0 & 71.0 & 57.7 & 65.0 & 66.0 \\
        & X-Decoder~\cite{zou2023generalized} \textcolor{gray}{(CVPR-23)}  & - & - & - & - & - & - & 64.6 & - \\
        & SEEM~\cite{zou2023segment} \textcolor{gray}{(arXiv-23)}  & - & - & - & - & - & - & 65.7 & -   \\
        & LISA~\cite{lai2023lisa} \textcolor{gray}{(arXiv-23)}   & 74.9 & 79.1 & 72.3 & 65.1 & 70.8 & 58.1 & 67.9 & 70.6  \\
        \cline{2-10}
        \rowcolor{mygray} \cellcolor{white} & \textbf{UniRES} (Ours) & \textbf{77.4} & \textbf{80.9} & \textbf{74.7} & \textbf{69.4} & \textbf{76.1} & \textbf{61.4} & \textbf{69.0} & \textbf{71.7} \\
        \midrule
        \multirow{6}{*}{\rotatebox{90}{mIoU}} & VLT \cite{ding2021vision} \textcolor{gray}{(ICCV-21)} & 65.7 & 68.3 & 62.7 & 55.5 & 59.2 & 49.4 & 53.0 & 56.7 \\
        & RefTr \cite{li2021referring}  \textcolor{gray}{(NeurIPS-21)} & 74.3 & 76.8 & 70.9 & 66.8 & 70.6 & 59.4 & 66.6 & 67.4 \\
        & SeqTR \cite{zhu2022seqtr} \textcolor{gray}{(ECCV-22)} & 71.7 & 73.3 & 69.8 & 63.0 & 66.7 & 59.0 & 65.0 & 65.7 \\
        & CRIS \cite{wang2022cris} \textcolor{gray}{(CVPR-22)} & 70.5 & 73.2 & 66.1 & 62.3 & 68.1 & 53.7 & 59.9 & 60.4 \\
        & LAVT \cite{yang2022lavt} \textcolor{gray}{(CVPR-22)}  & 74.5 & 76.9 & 70.9 & 65.8 & 71.0 & 59.2 & 63.3 & 63.6 \\
        \cline{2-10}
        \rowcolor{mygray} \cellcolor{white} & \textbf{UniRES} (Ours) & \textbf{79.2} & \textbf{81.6} & \textbf{76.6} & \textbf{73.0} & \textbf{78.1} & \textbf{65.8} & \textbf{71.7} & \textbf{73.2} \\
        \specialrule{.1em}{.05em}{.05em}
    \end{tabular}
    \label{tab:sota}}
    \end{center}
    \vspace{-20pt}
\end{table*}

\noindent \textbf{Comparison with SOTA Methods.} To validate the superiority of our MRES-32M dataset and model UniRES, our framework is fairly evaluated against previous SOTA methods on RefCOCO \cite{yu2016modeling}, RefCOCO+ \cite{yu2016modeling} and RefCOCOg \cite{nagaraja2016modeling} under both zero-shot and fine-tuning settings. 
As presented in Table \ref{tab:sota}, our method greatly outperforms previous methods in terms of segmentation accuracy across all the benchmark datasets. 
Via directly zero-shot transferring to the downstream classic RES task after pre-training on our MRES-32M dataset, Our UniRES model achieves a leading zero-shot segmentation accuracy (\ie, approximately 71\%) without any fine-tuning to adapt to downstream task's data, which is significantly higher than all the recently proposed zero-shot RES methods by a large margin (\ie, ↑30-40\% mIoU).
At the same time, it is worth noting that our zero-shot segmentation performance is already better than many of previous RES methods under fine-tuning setting (\eg, CRIS \cite{wang2022cris} and ReSTR \cite{kim2022restr}), validating the potential of our MRES-32M dataset and UniRES model.
Furthermore, after fine-tuning on the classic RES datasets, our UniRES greatly surpasses previous methods no matter the specialist models (\eg, LAVT \cite{yang2022lavt} and LISA \cite{lai2023lisa}) for classic RES or the generalist models (\eg, X-Decoder \cite{zou2023generalized} and SEEM \cite{zou2023segment}).

\subsection{Ablation Studies}
\label{ablation}
We conduct ablation experiments on our RefCOCOm validation set. 
\textit{The tables below involves three evaluation settings on our RefCOCOm, using only objects as RES targets, using only parts as RES targets, and a mixed setting that combines both of the above two granularity levels.}

\vspace{-3mm}
\subsubsection{Ablation Study on MRES-32M Dataset}

\noindent \textbf{Data Granularity.} We first probe into the effect of pre-training data's granularity with 50\% of our MRES-32M dataset. 
As shown in Table \ref{tab:Data_Granularity}, the baseline without pre-training on MRES-32M dataset obtains 75.2\%, 15.8\% and 32.0\% mIoU separately on our RefCOCOm validation set under three granularity levels. 
Either introducing the object-level or part-level data from MRES-32M dataset for pre-training consistently results in an considerable accuracy increase across different granularity settings.
In fact, pre-training on data at a certain granularity level to improve performance on corresponding benchmark aligns with common sense. 
However, it is noteworthy that incorporating part-level data into training also enhances model performance on the only object-level RefCOCOm. 
This underscores the significance of part-level understanding, as emphasized in the Sec. \ref{introduction}, where grasping the nuances of parts can yield benefits for object-level grounding.
Besides, by jointly incorporating the training samples of both granularity, our method  attains 2.8\% improvement against baseline under all the granularity settings on RefCOCOm,
which fully demonstrate the benefit of exploiting our MRES-32M dataset for both object-level and part-level RES tasks. 
\vspace{-3mm}
\begin{table}[htbp]
    \small
    \setlength{\belowcaptionskip}{1.0pt}
    \begin{center}
    \caption{Ablation study on the granularity of pre-training data in our proposed MRES-32M. Object and Part denote the introduction of object-level and part-level data for pre-training.}
     \vspace{-3mm}
     \setlength{\tabcolsep}{1.2mm}{\begin{tabular}{c|c|ccc}
      \specialrule{.1em}{.05em}{.05em} 
      \multirow{2}{*}{Object} & \multirow{2}{*}{Part} & \multicolumn{3}{c}{RefCOCOm}\\
         &  & Object-Only & Part-Only & Object\!\;\&\;\!Part  \\
        \hline
         &  & 75.2 & 15.8 & 30.5 \\ 
        \ding{51} &  & 77.5 & 15.9 & 31.1  \\
         & \ding{51} & 75.9  & 18.4  & 32.6  \\
        \rowcolor{mygray} \ding{51} & \ding{51}  & \textbf{78.0} & \textbf{18.6} & \textbf{33.3} \\
    \specialrule{.1em}{.05em}{.05em}
    \end{tabular}
    \vspace{-5mm}
    \label{tab:Data_Granularity}}
    \end{center}
\end{table}

\vspace{-3mm}
\begin{table}[htbp]
    \small
    \setlength{\belowcaptionskip}{1.0pt}
    \begin{center}
    \caption{Ablation study on the data scale of MRES-32M dataset.}
     \vspace{-3mm}
     \setlength{\tabcolsep}{1.2mm}{\begin{tabular}{c|ccc}
      \specialrule{.1em}{.05em}{.05em} 
      \multirow{2}{*}{Ratios} & \multicolumn{3}{c}{RefCOCOm}\\
       &  Object-Only & Part-Only & Object\!\;\&\;\!Part  \\
        \hline
        0\%  & 75.2  &  15.8 & 30.5   \\ 
        20\% & 76.8  &  17.3 & 32.0   \\ 
        50\% & 78.0  &  18.6 & 33.3   \\
        \rowcolor{mygray} 100\% & \textbf{79.2} &  \textbf{19.6} & \textbf{34.3} \\
    \specialrule{.1em}{.05em}{.05em}
    \end{tabular}
    \vspace{-8mm}
    \label{tab:Data_Scale}}
    \end{center}
\end{table}

\noindent \textbf{Data Scale.}
Next, we investigate the effect of different percentages of training samples in our MRES-32M dataset. 
The results are presented in Table \ref{tab:Data_Scale}. 
It is obvious in Table \ref{tab:Data_Scale} that the model performance for our MRES tasks is consistently improved with more and more employed training samples, which implicitly validates the high quality of our built dataset.
As the ratios of introduced training samples continue to rise, there's no sign of diminishing gains in model performance, suggesting that our framework has significant potential with continually scaled up training data.

\vspace{-3mm}
\begin{table}[htbp]
    \small
    \setlength{\belowcaptionskip}{1.0pt}
    \begin{center}
    \caption{Ablation study on the effectiveness of our MRES-32M dataset on RES SOTA methods. }
    \vspace{-3mm}
    \setlength{\tabcolsep}{0.8mm}{\begin{tabular}{l|c|ccc}
    \specialrule{.1em}{.05em}{.05em} 
    \multirow{2}{*}{Methods} & \multirow{2}{*}{MRES-32M} & \multicolumn{3}{c}{RefCOCOm} \\
      &   &  Object-Only & Part-Only & Object\!\;\&\;\!Part  \\
    \hline
    CRIS  &   & 70.5   & 10.6   & 25.4   \\ 
    \rowcolor{mygray} CRIS & \ding{51}  & \textbf{73.1} & \textbf{15.5}  & \textbf{29.7} \\ 
    \hline
    LAVT &    & 74.5 &  15.3 & 29.9  \\ 
    \rowcolor{mygray} LAVT  & \ding{51} & \textbf{75.7} & \textbf{19.3} & \textbf{33.2} \\
    \specialrule{.1em}{.05em}{.05em}
    \end{tabular}
    \vspace{-7mm}
    \label{tab:onothermethods}}
    \end{center}
\end{table}

\noindent \textbf{Dataset Necessity.}
To prove the necessity and effectiveness of our MRES-32M dataset for the proposed MRES task, we take previous SOTA methods (\ie, CRIS \cite{wang2022cris} and LAVT \cite{yang2022lavt}) on classic RES and compare the segmentation accuracy of the same models with or without pre-training on our MRES-32M.
As shown in Table \ref{tab:onothermethods}, the original CRIS and LAVT can already well handle the classic RES task with only object-level grounding skills, but it performs poorly on the multi-grained RefCOCOm benchmark, which requires part-level grounding capability.
In contrast, pre-training on our MRES-32M consistently leads to great performance gains no matter on object-only, part-only or the multi-granularity RES tasks, because our high-quality MRES-32M dataset can effectively enable the model to handle the part-level visual grounding task and enhances the original object-level grounding capability.

\vspace{-3mm}
\subsubsection{Ablation Study on UniRES Model}

\vspace{-5mm}
\begin{table}[htbp]
    \small
    \setlength{\belowcaptionskip}{1.0pt}
    \begin{center}
    \caption{Ablation study on the query-based grouping design. }
    \vspace{-3mm}
     \setlength{\tabcolsep}{0.8mm}{\begin{tabular}{c|c|ccc}
      \specialrule{.1em}{.05em}{.05em} 
      \multirow{2}{*}{High-Level} & \multirow{2}{*}{Low-Level} & \multicolumn{3}{c}{RefCOCOm}\\
         &  & Object-Only & Part-Only & Object\!\;\&\;\!Part  \\
        \hline
        &  & 74.4 & 14.9 & 29.6  \\ 
        \ding{51} &  & 74.9 & 15.2 & 30.0  \\
        & \ding{51}  & 74.9 & 15.4 & 30.1  \\
        \rowcolor{mygray} \ding{51} & \ding{51} & \textbf{75.2}  & \textbf{15.8} & \textbf{30.5} \\
    \specialrule{.1em}{.05em}{.05em}
    \end{tabular}
    \vspace{-8mm}
    \label{tab:Model_Structure}}
    \end{center}
\end{table}

\begin{figure}[htbp]
    \centering
    \includegraphics[width=0.42\textwidth]{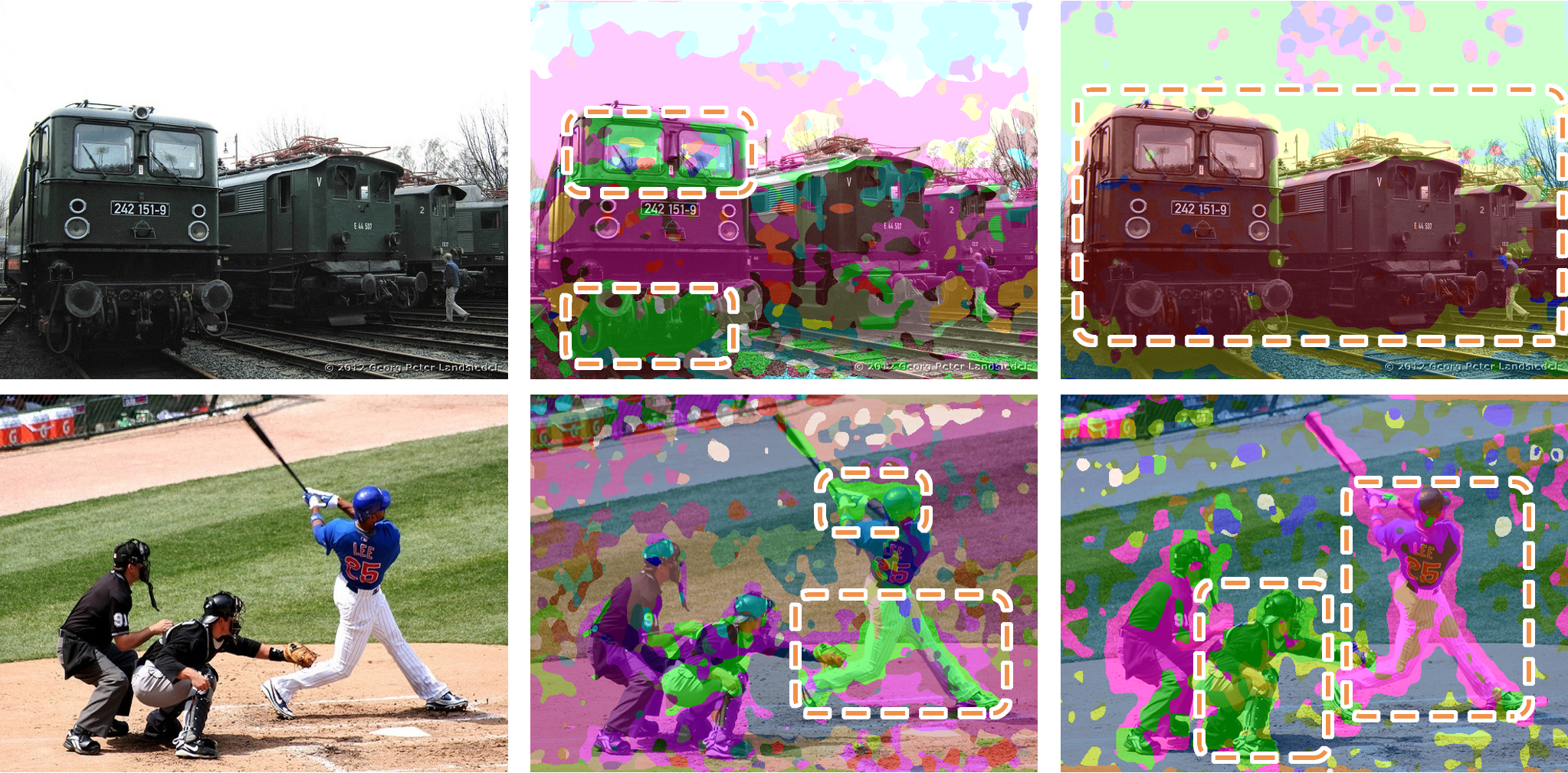}
    \begin{tabu} to 0.88 \linewidth{X[1.0c] X[1.0c] X[1.0c]} 
        \scriptsize{(a) Image} &  \scriptsize{(b) Low-Level} &  \scriptsize{(c) High-Level} \\
    \end{tabu}
    \vspace{-10pt}
    \caption{Qualitative analysis for ablation study on the object-level and part-level grouping design in our model structure. (a) the input image. (b) low-level group tokens. (c) high-level group tokens.}
    \label{fig_grouptokens_vis}
    \vspace{-6mm}
\end{figure}

Besides, we additionally conduct ablation studies on the structural design of our model UniRES.
As presented in Table \ref{tab:Model_Structure}, removing the simple but effective grouping design at two different levels  (this simultaneously leads to discarding the second-stage decoder) will result in a considerable deterioration in model performance. 
Since the sequentially appended pixel grouping tokens are introduced to effectively capture the high-level and low-level visual features for further visual feature enhancement, losing any type of these group tokens will lead to a decrease in the model's segmentation accuracy on both object-level and part-level RES tasks. In order to verify that the high-level and low-level pixel grouping tokens can capture the clustering features of the corresponding level, we also visualize the group tokens of the two levels (before being sent to LRF) for qualitative analysis, where the different colors represent the various areas of the same clustering group.
The visualization results in Fig. \ref{fig_grouptokens_vis} affirms that our introduced low-level and high-level group tokens respectively capture the part-level local features and aggregate the object-level global features with stronger semantics through the pixel grouping process, which aligns with our intentions for the grouping design.

\vspace{-2mm}

\section{Conclusion and Broader Impact}

In this paper, we move beyond previous works that focused solely on object-level visual grounding tasks and take a step further to finer-grained part-level RES. 
We put forward a new multi-granularity referring expression segmentation task and establish an evaluation benchmark named RefCOCOm by manual annotation.
To advance the visual grounding at both object and part levels towards finer-grained vision-language understanding, we build the largest visual grounding dataset MG-32M to date, which is also the first dataset to provide part-level vision-language annotations. 
Furthermore, we have developed a simple yet strong multi-grained referring segmentation model called UniRES. 
As a baseline for our newly proposed MRES task, UniRES achieves the new state-of-the-art performance on both our RefCOCOm for MRES task and three classic RES datasets.
We plan to release our RefCOCOm benchmark, the MG-32M dataset, and the UniRES model to the public, aspiring to foster future research in fine-grained visual grounding tasks and to inspire new research in this direction.

\section*{Acknowledgement}

We thank Yepeng Tang for the helpful discussions on this work, all the Image \& Video Analysis Group (IVA)'s members in CASIA for the technical support, and all the insightful reviewers for the helpful suggestions.
This work was supported by the National Science and Technology Major Project (No.2022ZD0118801), National Natural Science Foundation of China (U21B2043, 62206279).

{
    \small
    \bibliographystyle{ieeenat_fullname}
    \bibliography{main}
}

\clearpage
\setcounter{page}{1}

\appendix
\section{Appendix}
In this appendix section, we provide following items:
\begin{itemize}[noitemsep,leftmargin=*]
    \item (Sec. \textcolor{red}{1}) Potential limitations and future works.
    \item (Sec. \textcolor{red}{2}) Detailed information about the three benchmark datasets (\ie RefCOCO, RefCOCO+ and RefCOCO+) for classic referring expression segmentation (RES) task.
    \item (Sec. \textcolor{red}{3}) Implementation details on the three benchmark datasets (\ie RefCOCO, RefCOCO+ and RefCOCO+) for classic RES task and the newly built RefCOCOm benchmark for our multi-granularity RES (MRES) task.
    \item (Sec. \textcolor{red}{4}) More quantitative analysis about our newly built MRES-32M dataset.
    \item (Sec. \textcolor{red}{5}) More visualization results of visual comparisons, and the selected samples from our newly built RefCOCOm benchmark and MRES-32M dataset.
\end{itemize}

\subsection{Limitation and Future Work}

One potential limitation of this work is that the data scale of our MRES-32M and model capacity of UniRES could be scaled up further to push SOTA performance. 
Moreover, in this work we mainly focus on the (M)RES task with visual masks as output, which means that although our framework demonstrates the new crucial part-level referring segmentation capabilities, it currently can not produce textual responses and thus can not handle the tasks related to vision-language conversations. 
However, the integration of large language models could enhance our framework's text comprehension and generation abilities, allowing it to be accordingly adjusted to overcome this limitation.
This opens up a future research direction to develop a more general and powerful framework based on multimodal large language models that could interact with user-provided textual and visual inputs across multiple levels of granularity.

\subsection{Details about the Benchmark Datasets}

\textbf{RefCOCO} \cite{yu2016modeling}, stands as one of the largest and frequently utilized datasets derived from MSCOCO \cite{lin2014microsoft} for the task of referring expression segmentation. It comprises 142,209 annotated expressions with an average expression length of 3.6 words, labeling 50,000 objects across 19,994 images. The dataset is partitioned into 120,624 training samples, 10,834 validation samples, and two test subsets—test A and test B—containing 5,657 and 5,095 instances, respectively.

\textbf{RefCOCO+} \cite{yu2016modeling} encompasses 141,564 language expressions with a slightly shorter average expression length of 3.5 words, targeting 49,856 objects within 19,992 images. This dataset follows a similar split as RefCOCO, offering 120,624 training, 10,758 validation, 5,726 test A, and 4,889 test B samples. Unique to RefCOCO+, it omits expressions that use absolute location terms, posing an increased challenge for the classic RES task.

\textbf{RefCOCOg} \cite{nagaraja2016modeling}, serves as the third benchmark dataset and contains 104,560 referring expressions,  with an significantly longer average length of 8.4 words for  54,822 objects in 26,711 images. The language expressions in this dataset is sourced from Amazon Mechanical Turk, marking a distinction from the previous two datasets. Following previous works, we employ the UMD partition standard \cite{hu2016segmentation} for our evaluations in this paper.

\subsection{Implementation Details}

\noindent \textbf{Experimental Setup.} 
Our work is implemented based on Pytorch \cite{paszke2019pytorch} and trained with NVIDIA A800 GPUs. 
Considering the crucial scalability and the ease of implementation, the Vision Transformer \cite{dosovitskiy2020image} is adopted as the image encoder for all the experiments.
The text and image encoder are initialized by CLIP \cite{radford2021learning}, while the rest part of model weights are randomly initialized. 
During training, 
with 128 and 64 batch size respectively, the AdamW optimizer and a weight decay of 5e-4 are adopted to pre-train and fine-tune our model for 50 and 15 epochs. 
With a warm-up strategy for 5-epoch pre-training on our MRES-32M and 1-epoch fine-tuning on the specific downstream grounding dataset, the initial learning rate is set to 1e-5 with a cosine decay schedule.
Following CRIS \cite{wang2022cris}, due to the extra [SOS] and [EOS] tokens, the input sentences are set with a maximum sentence length of 17 for RefCOCO, our RefCOCOm and RefCOCO+, 22 for RefCOCOg. 

During inference, the predicted results by our method is upsampled back to the original image size and binarized with a threshold of 0.35 to the final segmentation result. 
Any extra post-processing operations or inference tricks can be exploited to further boost the segmentation accuracy but are not included in this work.

\subsection{More Analysis about MRES-32M Dataset}
\begin{figure*}[htbp]
    \centering
    \includegraphics[width=0.98\textwidth]{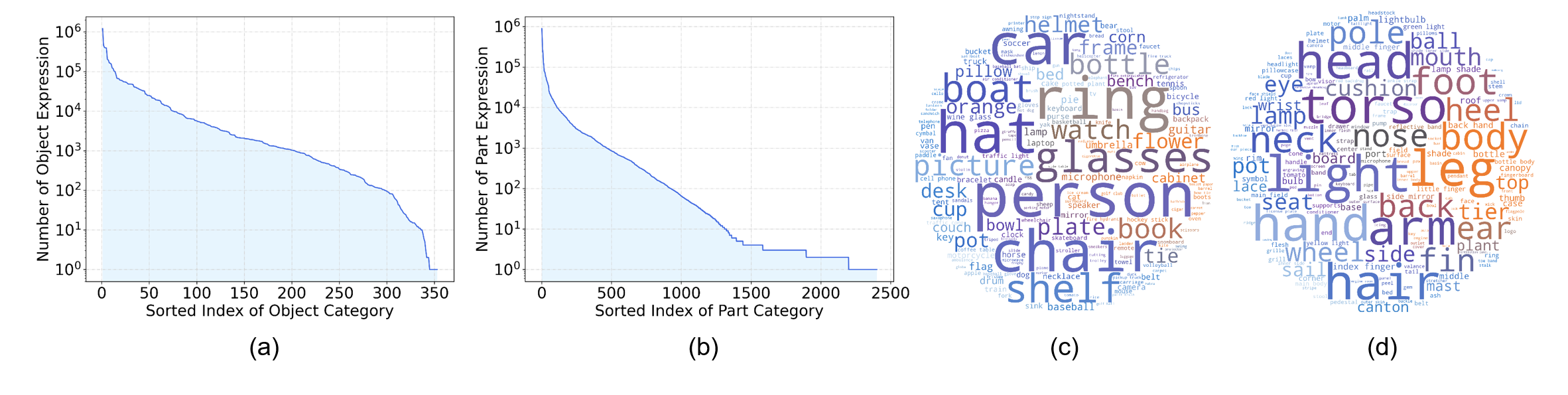}
    \vspace{-15pt}
    \caption{
    MRES-32M dataset statistics.
    (a) the number of referring expressions per objects' category in the log scale. 
    (b) the number of referring expressions per parts' category in the log scale. 
    (c) the word cloud highlights the head objects' categories. 
    (d) the word cloud highlights the head parts' categories. 
    }
    \label{fig_analysisofMRES-32M}
    \vspace{-3mm}
\end{figure*}

Our MRES-32M is composed of 365 object categories and an associated 2,299 part categories. Compared with existing datasets, it covers a wider range of multimodal knowledge and is an important step towards open-world understanding. Among our MRES-32M dataset, the number of referring expressions per objects’ category and per parts’ category, the word cloud that highlights the head objects' and parts' categories are both presented in Fig. \ref{fig_analysisofMRES-32M}. 

\subsection{More Visualization Results}
\noindent \textbf{Visual Comparison with SOTA Methods.}
\begin{figure*}[htbp]
    \centering
    \includegraphics[width=0.80\textwidth]{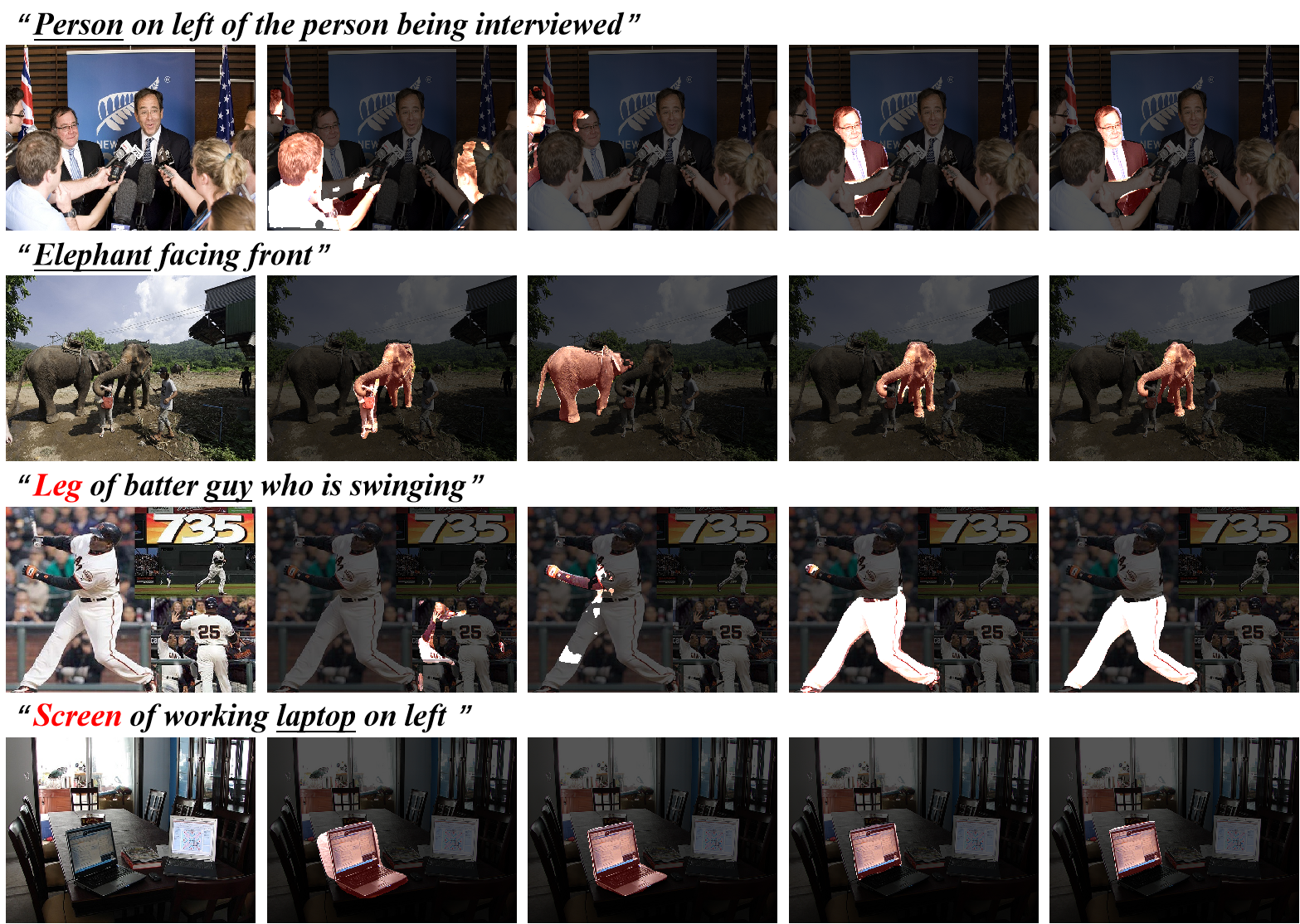}
    \begin{tabu} to 0.80\linewidth{X[1.0c] X[1.0c] X[1.0c] X[1.0c] X[1.0c]} 
        \scriptsize{(a) Image} &  \scriptsize{(b) CRIS} &  \scriptsize{(c) LAVT}  &  \scriptsize{\textbf{(e) Ours}} &  \scriptsize{(d) GT} \\
    \end{tabu}
    \vspace{-5pt}
    \caption{The visual comparison of segmentation results on our RefCOCOm validation set. (a) the input image. (b) CRIS. (c) LAVT. (d) our UniRES. (e) the ground truth. }
    \label{fig_SOTAComparison_vis}
    \vspace{-10pt}
\end{figure*}
In addition, to validate the segmentation quality of our framework, classic RES methods CRIS \cite{wang2022cris} and LAVT \cite{yang2022lavt}, 
as well as our proposed UniRES are further adopted for qualitative comparison. 
The provided visualization results in the first and second row of Fig.~\ref{fig_SOTAComparison_vis} convinces that our method can better accomplish the classic object-level RES task and generate much better fine-grained segmentation masks of the target objects, greatly reducing the over-segmentation and under-segmentation errors.
Similarly, as shown in the third and fourth row in Fig.~\ref{fig_SOTAComparison_vis}, when facing the more challenging part-level grounding task, our approach can clearly locate and segment the referring target regions more accurately while the other previous methods fail to reach the same level.

\begin{figure*}[htbp]
    \centering
    \includegraphics[width=0.75\textwidth]{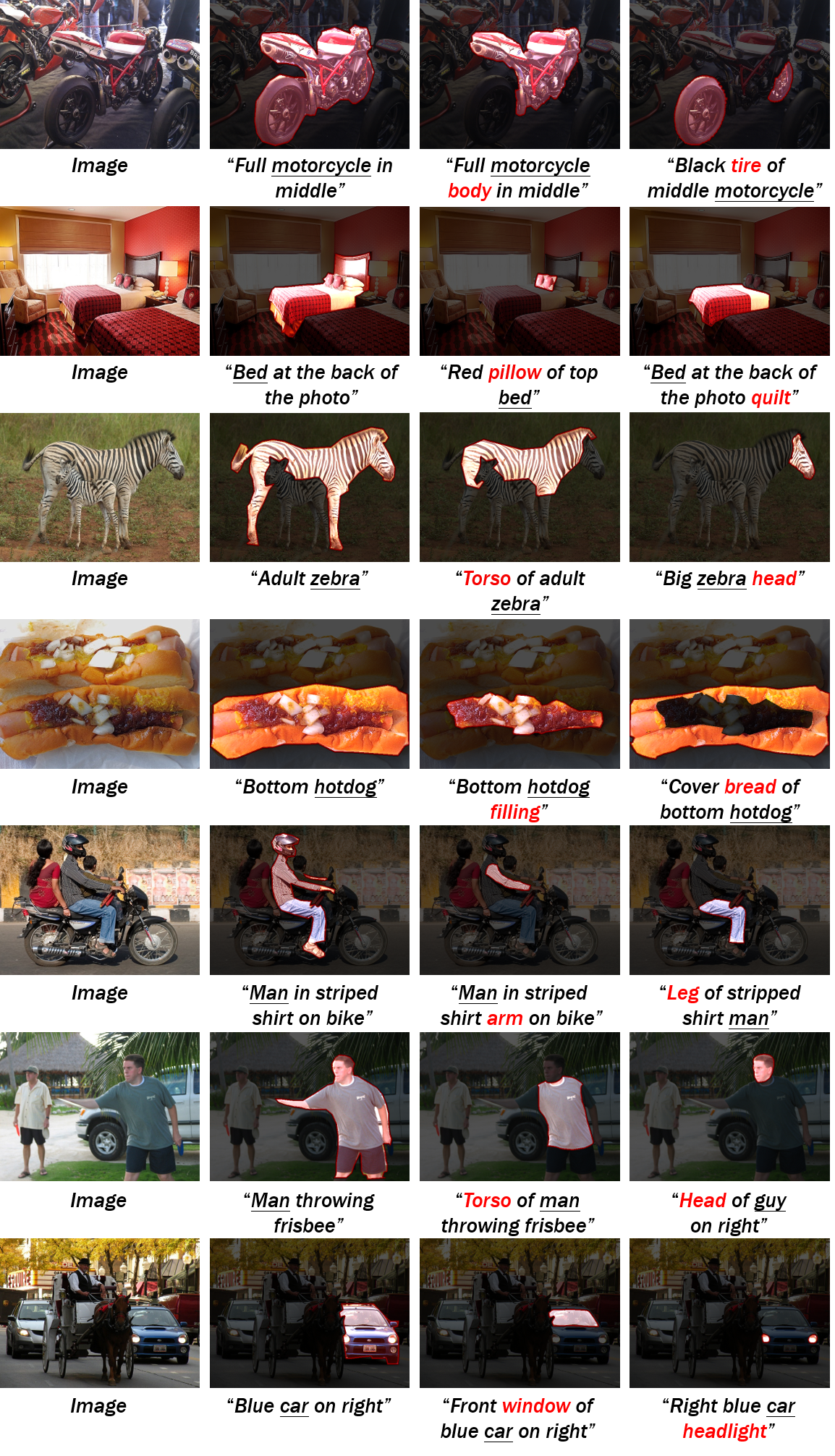}
    \vspace{-10pt}
    \caption{
    More selected samples from our proposed RefCOCOm benchmark for multi-granularity RES task. 
    }
    \label{Supplementary_fig_examplesofRefCOCOm}
\end{figure*}

\noindent \textbf{Samples of RefCOCOm Benchmark.} 
We have also provided a few more examples in our RefCOCOm benchmark for proposed MRES task in Fig. \ref{Supplementary_fig_examplesofRefCOCOm}.

\begin{figure*}[htbp]
    \centering
    \includegraphics[width=0.68\textwidth]{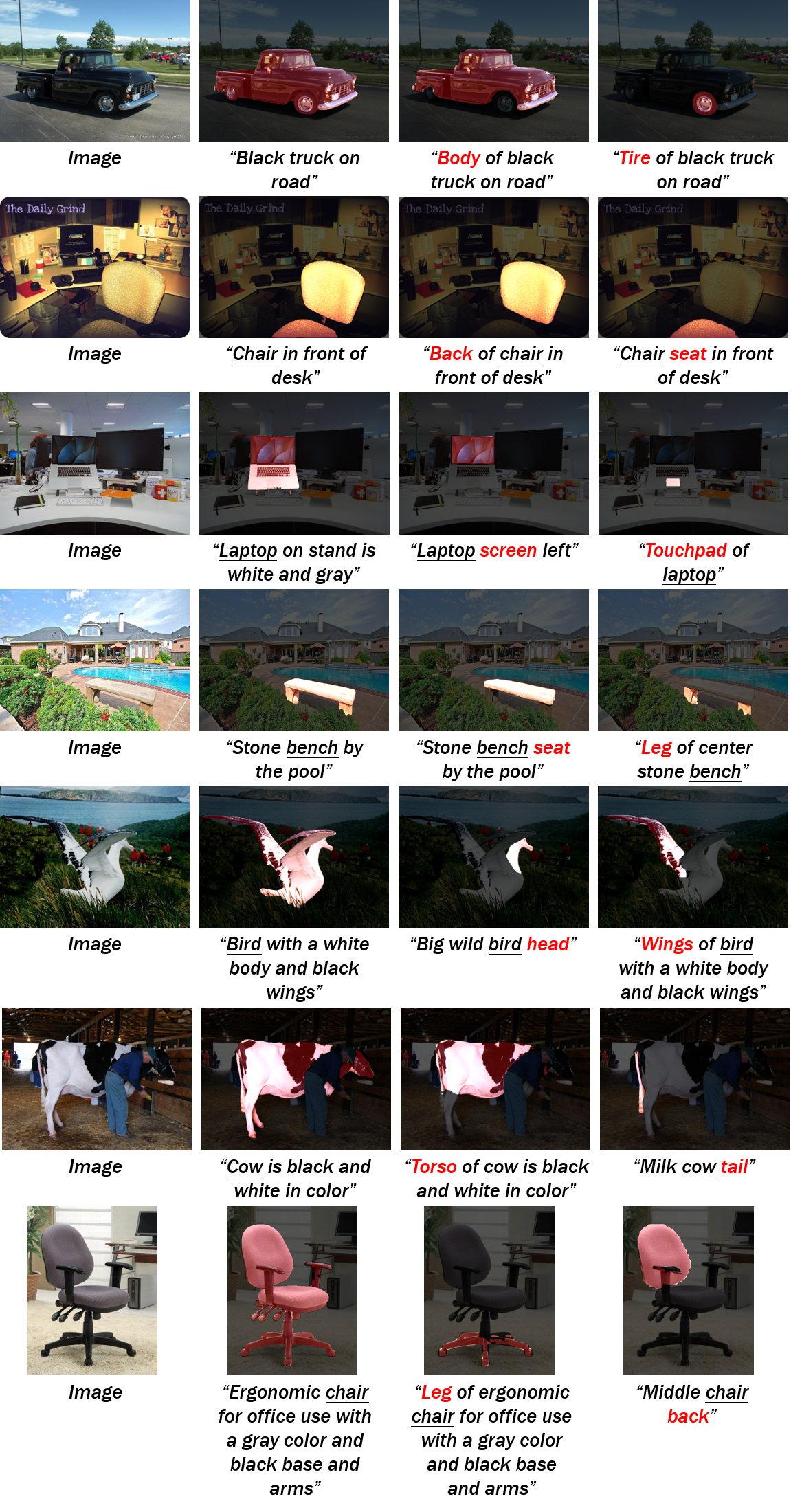}
    \vspace{-10pt}
    \caption{
    Selected samples from our built MRES-32M dataset for visual grounding tasks. 
    }
    \label{Supplementary_fig_examplesofMRES-32M}
\end{figure*}

\noindent \textbf{Samples of MRES-32M Dataset.} 
A few examples in our newly built MRES-32M dataset for visual grounding task are presented in Fig. \ref{Supplementary_fig_examplesofMRES-32M}. 


\end{document}